\documentclass{article}

\pdfoutput=1

\usepackage{microtype}
\usepackage{amsmath}
\usepackage{amssymb}
\usepackage{mathtools}
\usepackage{amsthm}
\usepackage{comment}

     \usepackage[final]{neurips_2023}

\usepackage[utf8]{inputenc} 
\usepackage[T1]{fontenc}    
\usepackage{hyperref}       
\usepackage{url}            
\usepackage{booktabs}       
\usepackage{amsfonts}       
\usepackage{nicefrac}       
\usepackage{microtype}      
\usepackage{xcolor}         
\usepackage{tabularx}
\newcolumntype{Y}{>{\centering\arraybackslash}X}
\usepackage{graphicx}

\author{%
  Wenhao Wang \thanks{Work done during Wenhao Wang’s internship at Baidu Inc.}\\
  ReLER, University of Technology Sydney
  \And
  Yifan Sun\\
  Baidu Inc.
  \AND
  Wei Li\\
  Zhejiang University
  \And
  Yi Yang \thanks{Corresponding author.}\\
  Zhejiang University
}

\begin{document}
\title{TransHP: Image Classification with Hierarchical Prompting}

\maketitle

\begin{abstract}
This paper explores a hierarchical prompting mechanism for the hierarchical image classification (HIC) task. 
Different from prior HIC methods, our hierarchical prompting is the first to explicitly inject ancestor-class information as a tokenized hint that benefits the descendant-class discrimination. We think it well imitates human visual recognition, \emph{i.e.}, humans may use the ancestor class as a prompt to draw focus on the subtle differences among descendant classes. We model this prompting mechanism into a Transformer with Hierarchical Prompting (TransHP). TransHP consists of three steps: 1) learning a set of prompt tokens to represent the coarse (ancestor) classes, 2) on-the-fly predicting the coarse class of the input image at an intermediate block, and 3) injecting the prompt token of the predicted coarse class into the intermediate feature.
Though the parameters of TransHP maintain the same for all input images, the injected coarse-class prompt conditions (modifies) the subsequent feature extraction and encourages a dynamic focus on relatively subtle differences among the descendant classes. 
Extensive experiments show that TransHP improves image classification on accuracy (\emph{e.g.}, improving ViT-B/16 by $+2.83\%$ ImageNet classification accuracy), training data efficiency (\emph{e.g.}, $+12.69\%$ improvement under $10\%$ ImageNet training data), and model explainability. Moreover, TransHP also performs favorably against prior HIC methods, showing that TransHP well exploits the hierarchical information. The code is available at: https://github.com/WangWenhao0716/TransHP.
\end{abstract}

\section{Introduction}
Hierarchical image classification (HIC) aims to exploit the semantic hierarchy to improve prediction accuracy. More concretely, HIC provides additional coarse labels (\emph{e.g.}, Rose) which indicate the ancestors of the relatively fine labels (\emph{e.g.}, China Rose and Rose Peace). The coarse labels usually do not need manual annotation and can be automatically generated based on the fine labels, \emph{e.g.}, through WordNet \cite{miller1998wordnet} or word embeddings \cite{mikolov2013efficient}. Since it barely increases any annotation cost while bringing substantial benefit, HIC is of realistic value and has drawn great research interest \cite{landrieu2021leveraging, zhang2022use}.  

This paper explores a novel hierarchical prompting mechanism that well imitates the human visual recognition for HIC. Specifically, a person may confuse two close visual concepts (\emph{e.g.}, China Rose and Rose Peace) when the scope of interest is large (\emph{e.g.}, the whole Plantae). However, given a prompt narrowing down the category range (\emph{e.g.}, the rose family), the person can shift his/her focus to the subtle variations within the coarse class. 
We duplicate this procedure for deep visual recognition based on the transformer prompting technique. The transformer prompting typically uses prompts (implemented as tokens or vectors) to adapt a pre-trained transformer for different downstream tasks \cite{li2021prefix,gu2021ppt,he2021towards}, domains \cite{ge2022domain}, \emph{etc}. In this paper, we inject coarse-class prompt into
the intermediate stage of a transformer. The injected coarse-class prompt will then modify the following feature extraction for this specific coarse class, yielding the so-called hierarchical prompting. 
To the best of our knowledge, explicitly injecting the coarse class information as a prompt has never been explored in the HIC community.

\begin{figure*}[t] 
\centering 
\includegraphics[width=1\textwidth]{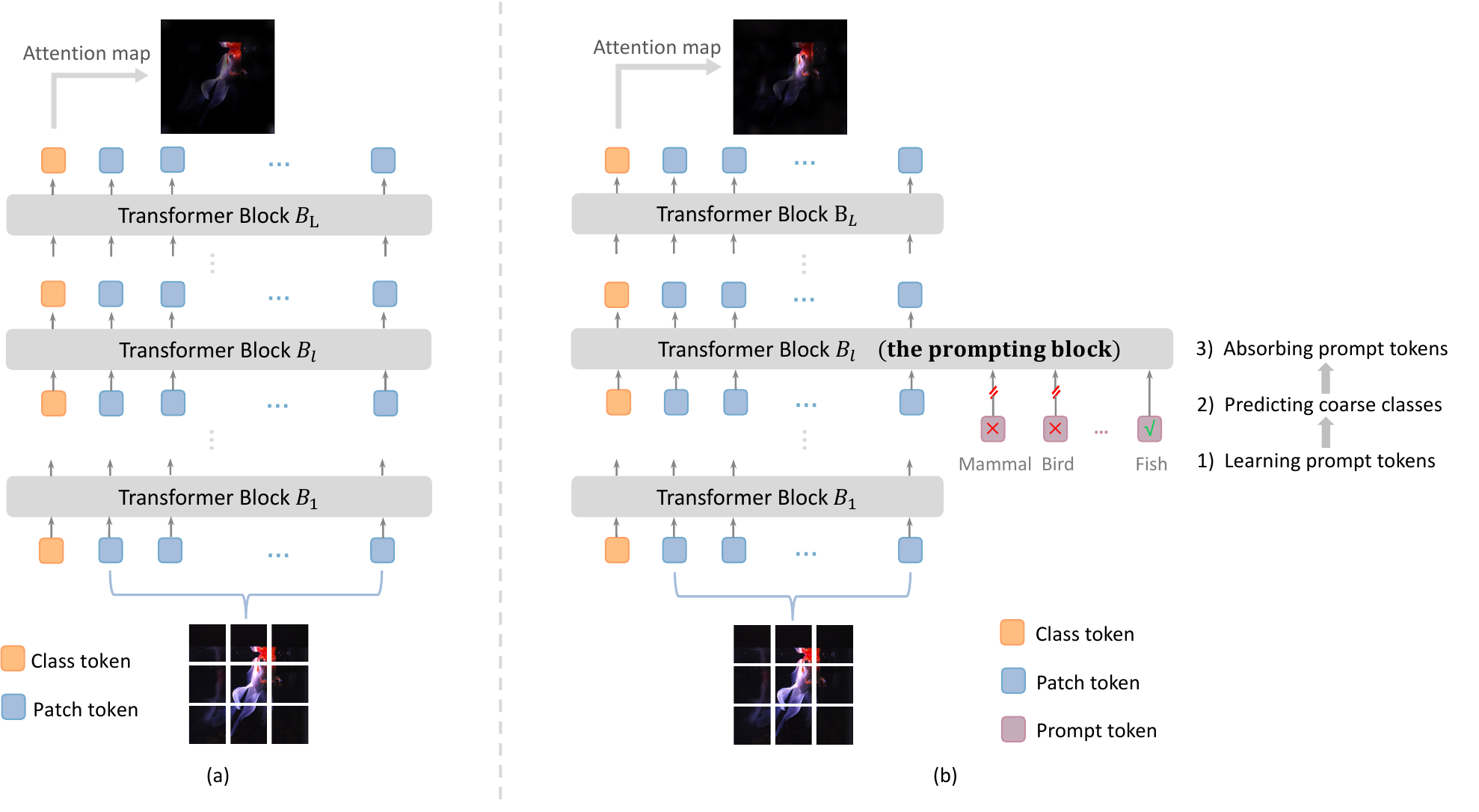} 
\caption{The comparison between Vision Transformer (ViT) and the proposed Transformer with Hierarchical Prompting (TransHP). In (a), ViT attends to the overall foreground region and recognizes the goldfish from the 1000 classes in ImageNet. In (b), TransHP uses an intermediate block to recognize the input image as belonging to the fish family and then injects the corresponding prompt. Afterward, the last block attends to the face and crown, which are particularly informative for distinguishing the goldfish against other fish species. Please refer to Fig.~\ref{Fig: subtle} for more visualizations. Note that TransHP may have multiple prompting blocks corresponding to multi-level hierarchy.}
\label{Fig: compare} 
\vspace*{-2mm}
\end{figure*}

We model our hierarchical prompting mechanism into a Transformer with Hierarchical Prompting (TransHP). Fig.~\ref{Fig: compare} compares our TransHP against a popular transformer backbone ViT \cite{dosovitskiy2021an}. 
TransHP consists of three steps: 
1) TransHP learns a set of prompt tokens to represent all the coarse classes and selects an intermediate block as the ``prompting block'' for injecting the prompts. 
2) The prompting block on-the-fly predicts the coarse class of the input image.
3) The prompting block injects the prompt token of the predicted class (\emph{i.e.}, the target prompt token) into the intermediate feature. 
Specifically, TransHP concatenates the prompt tokens with feature tokens (\emph{i.e.}, the ``class'' token and the patch tokens) from the preceding block, and then feeds them into the prompting block, where the feature tokens absorb information from the target prompt through cross-attention \footnote{In practice, the absorption is in a ``soft'' manner which assigns all the prompt tokens with soft weights.}. 

Although TransHP is based on the prompting mechanism of the transformer, it has fundamental differences against prior transformer prompting techniques. 
A detailed comparison is in Section~\ref{sec: related}. 
We hypothesize this hierarchical prompting  will encourage TransHP to dynamically focus on the subtle differences among the descendant classes. Fig.~\ref{Fig: compare} validates our hypothesis by visualizing the attention map of the final-block class token. In Fig.~\ref{Fig: compare} (a), given a goldfish image, the baseline model (ViT) attends to the whole body for recognizing it from the entire 1000 classes in ImageNet. In contrast, in Fig.~\ref{Fig: compare} (b), since the intermediate block has already received the prompt of ``fish'', TransHP mainly attends to the face and crown which are particularly informative for distinguishing the goldfish against other fish species. Please refer to Section~\ref{sec: visualization} for more visualization examples. \par 

We conduct extensive experiments on multiple image classification datasets (\emph{e.g.}, ImageNet \cite{deng2009imagenet} and iNaturalist \cite{van2018inaturalist}) and show that the hierarchical prompting improves the accuracy, data efficiency and explainability of the transformer: \textbf{(1) Accuracy.} TransHP brings consistent improvement on multiple popular transformer backbones and five image classification datasets. For example, on ImageNet, TransHP improves ViT-B/16~\cite{dosovitskiy2021an} by +2.83\% top-1 accuracy. 
 \textbf{(2) Data efficiency.} While reducing the training data inevitably compromises the accuracy, TransHP maintains better resistance against the insufficient data problem. For example, when we reduce the training data of ImageNet to 10\%, TransHP enlarges its improvement over the baseline to +12.69\%. \textbf{(3) Explainability.} Through visualization, we observe that the proposed TransHP shares some similar patterns with human visual recognition \cite{johannesson2016visual,mirza2018human}, \emph{e.g.}, taking an overview for coarse recognition and then focusing on some critical local regions for the subsequent recognition after prompting.
 Moreover, TransHP also performs favorably against prior HIC methods, showing that TransHP well exploits the hierarchical information. 

\section{Related Works} \label{sec: related}

\textbf{Prior HIC methods} have never explored the prompting mechanism. We note that the prompting technique has not been introduced to the computer vision community until the very recent 2022~\cite{wang2022learning,wang2022dualprompt,luddecke2022image,zhang2022NOAH,jia2022vpt}. 
Two most recent and state-of-the-art HIC methods in 2022 are not related to the prompting technique either. Specifically, Guided \cite{landrieu2021leveraging} integrates a cost-matrix-defined metric into the supervision of a prototypical network. HiMulConE \cite{zhang2022use} builds an embedding space in which the distance between two classes is roughly consistent with the hierarchy (\emph{e.g.}, two sibling classes sharing the same ancestor are relatively close, and the classes with different ancestors are far away). Some earlier works \cite{su2021semi,jain2023test,yan2015hd} are valuable; however, they are also not directly related to the topic of prompting.\par

A deeper difference resulting from the prompting mechanism is how the mapping function of the deep model is learned. 
Specifically, a deep visual recognition model can be viewed as a mapping function from the raw image space into the label space. All these prior methods learn a shared mapping for all the images to be recognized. In contrast, the proposed TransHP uses the coarse-class prompt to condition itself (from an intermediate block). It can be viewed as specifying an individual mapping for different coarse classes, yielding a set of mapping functions. Importantly, TransHP makes all these mapping functions share the same transformer and conditions the single transformer into different mapping functions through the prompting mechanism.\par 

\textbf {Prompting} was first proposed in NLP tasks \cite{brown2020language,gao2021making,jiang2020can}, and then has drawn research interest from the computer vision community, e.g. continual learning \cite{wang2022learning,wang2022dualprompt}, image segmentation \cite{luddecke2022image}, and neural architecture search \cite{zhang2022NOAH}. VPT \cite{jia2022vpt} focuses on how to fine-tune pre-trained ViT models to downstream tasks efficiently. Prompting can efficiently adapt transformers to different tasks or domains while keeping the transform's parameters untouched. \par
Based on the prompting mechanism, our hierarchical prompting makes some \textbf{novel} explorations, \emph{w.r.t.} the prompting objective, prompting structure, prompt selection manner, and training process. \textbf{1) Objective:} previous methods usually prompt for different tasks or different domains. In contrast, TransHP prompts for coarse classes in the hierarchy, in analogy to the hierarchical prompting in human visual recognition. \textbf{2) Structure:} previous methods usually inject prompt tokens to condition the whole model. In contrast, in TransHP, the bottom blocks is completely shared, and the prompt tokens are injected into the intermediate blocks to condition the subsequent inference. Therefore, the prompting follows a hierarchical structure in accordance to the semantic hierarchy under consideration. \textbf{3) Prompt selection:} TransHP pre-pends all the prompt tokens for different coarse classes and  autonomously selects the prompt of interest, which is also new (as to be detailed in Section \ref{Sec: selection}). \textbf{4) Training process:} The prompting technique usually consists of two stages, \textit{i.e.}, pre-training a base model and then learning the prompts for novel downstream tasks. When learning the prompt, the pre-trained model is usually frozen. This pipeline is different from our end-to-end pipeline, \textit{i.e.} no more fine-tuning after this training.

\begin{figure*}[t] 
\centering 
\includegraphics[width=1\textwidth]{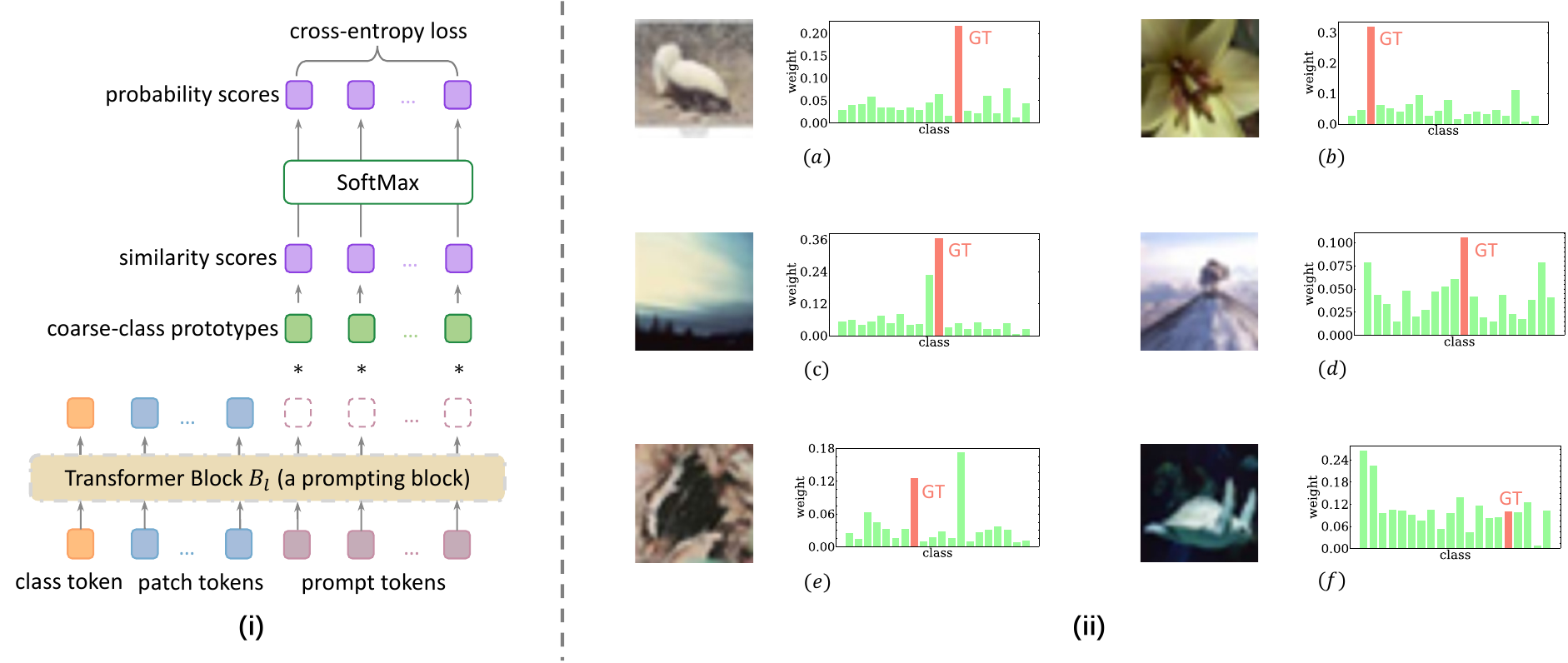} 
\vspace*{-2mm}
\caption{\textbf{(i) A prompting block in TransHP}. Instead of manually selecting the prompt of the coarse class, the prompting block pre-pends the whole prompt pool consisting of $M$ prompts ($M$ is the number of coarse classes) and performs autonomous selection. Specifically, it learns to predict the coarse class (Section~\ref{Sec: prompting block}) and spontaneously selects the corresponding prompt for absorption through soft weighting (Section~\ref{Sec: selection}), \emph{i.e.}, the predicted class has the largest absorption weight.
\textbf{(ii) Autonomous prompt selection.} We visualize the absorption weights of all the 20 coarse-class prompts for some CIFAR-100 images. 
It shows how TransHP selects the prompts when the coarse class prediction is correct ($a$ and $b$), ambiguous ($c$ and $d$), and incorrect ($e$ and $f$), respectively.  
The red and green columns correspond to the ground-truth (\textcolor{red}{GT}) class and the \textcolor{green}{false} classes, respectively.
The detailed investigation is in Section~\ref{Sec: selection}.}
\label{Fig: main} 
\end{figure*}
\section{Transformer with Hierarchical Prompting}
We first revisit a basic transformer for visual recognition (ViT~\cite{dosovitskiy2021an}) and the general prompting technique in Section~\ref{Sec: revisit}. Afterward, we illustrate how to reshape an intermediate block of the backbone into a hierarchical prompting block for TransHP in Section~\ref{Sec: prompting block}.
Finally, we investigate how the prompting layer absorbs the prompt tokens into the feature tokens in Section~\ref{Sec: selection}.

\subsection{Preliminaries}\label{Sec: revisit}
\textbf{Vision Transformer} (ViT) first splits an image into $N$ patches ($\left\{x_{i} \in \mathbb{R}^{3 \times P \times P} \mid i=1,2, \ldots, N\right\}$, where $P \times P$ is the patch size) and then embeds each patch into a $C$-dimensional embedding by $\mathbf{x}_{i}=\operatorname{Embed}\left(x_{i}\right)$. Afterward, ViT concatenates a class token $x^0_{cls}\in \mathbb{R}^{C}$ to the patch tokens and feed them into the stacked transformer blocks, which is formulated as:
\begin{equation}\label{equal: transformer}
\left[\mathbf{x}_{c l s}^{l}, \mathbf{X}^{l}\right]=B_{l}\left(\left[\mathbf{x}_{c l s}^{l-1}, \mathbf{X}^{l-1}\right]\right), \quad l=1,2, \ldots, L
\end{equation} 
where $\mathbf{x}_{c l s}^{l}$ and $\mathbf{X}^{l}$ are the class token and the patch tokens after the $l$-th transformer block $B_l$, respectively. After the total $L$ blocks, the final state of the class token ($\mathbf{x}_{c l s}^{L}$) is viewed as the deep representation of the input image and is used for class prediction. 
In this paper, we call the concatenation of class token and patch tokens (\emph{i.e.}, $\left[\mathbf{x}_{c l s}^{l-1}, \mathbf{X}^{l-1}\right]$) as the feature tokens.\par

\textbf{Prompting} was first introduced in Natural Language Processing to switch the same transformer model for different tasks by inserting a few hint words into the input sentences. More generally, it conditions the transformer to different tasks, different domains, \emph{etc}, without changing the transformer parameters but only changing the prompts. To condition the model for the $k$-th task (or domain), a popular practice is to select a prompt $\mathbf{p}_k$ from a prompt pool $\mathbf{P}=\{\mathbf{p}_0, \mathbf{p}_1, \cdots\}$ and pre-pend it to the first block. Correspondingly, Eqn.~\ref{equal: transformer} turns into:
\begin{equation}\label{equal: prompting}
\left[\mathbf{x}_{c l s}^{l}, \mathbf{X}^{l}, \mathbf{p}_k^{l}\right]=B_{l}\left(\left[\mathbf{x}_{c l s}^{l-1}, \mathbf{X}^{l-1}, \mathbf{p}_k^{l-1}\right]\right), 
\end{equation} 
where $\mathbf{p}_k \in \mathbf{P}$ (the superscript is omitted) conditions the transformer for the $k$-th task.

\subsection{The Prompting Block of TransHP}\label{Sec: prompting block}
The proposed TransHP selects an intermediate transformer block $B_l$ and reshapes it into a prompting block for injecting the coarse-class information.
Let us assume that there are $M$ coarse classes. Correspondingly, TransHP uses $M$ learnable prompt tokens $\mathbf{P}_{M}=\left[ \mathbf{p}_{0},\mathbf{p}_{1},...,\mathbf{p}_{M-1}\right]$  
to represent these coarse classes. 
Our intention is to inject $\mathbf{p}_k$ into the prompting layer, if the input image belongs to the $k$-th coarse class. \par

Instead of manually selecting the $k$-th prompt $\mathbf{p}_k$ (as in Eqn.~\ref{equal: prompting}), TransHP pre-pends the whole prompting pool $\mathbf{P}_{M}=\left[ \mathbf{p}_{0},\mathbf{p}_{1},...,\mathbf{p}_{M-1}\right]$ to the prompting layer and makes the prompting layer automatically select $\mathbf{p}_k$ for absorption. 
Specifically, through our design, TransHP learns to automatically 1) predict the coarse class, 2) select the corresponding prompt for absorption through ``soft weighting'', \emph{i.e.}, high absorption on the target prompt and low absorption on the non-target prompts. The learning procedure is illustrated in Fig.~\ref{Fig: main} (i). The output of the prompting layer is derived by:
\begin{equation} \label{equal: TransHP}
\left[\mathbf{x}_{c l s}^{l}, \mathbf{X}^{l}, \mathbf{\hat{P} }_{M} \right]=B_{l}\left(\left[\mathbf{x}_{c l s}^{l-1}, \mathbf{X}^{l-1}, \mathbf{P}_{M} \right]\right),
\end{equation}
where $\mathbf{\hat{P} }_{M}$ is the output state of the prompt pool $\mathbf{P}_{M}$ through the $l$-th transformer block $B_{l}$. $\mathbf{\hat{P} }_{M}$ will not be further forwarded into the following block. Instead, we use $\mathbf{\hat{P} }_{M}$ to predict the coarse classes of the input image. To this end, we compare $\mathbf{\hat{P} }_{M}$ against a set of coarse-class prototypes and derive the corresponding similarity scores by:
\begin{equation}
{S} = \{\mathbf{p}_i^\mathrm{T} \mathbf{w}_i\}, i=1,2,\cdots, M,
\end{equation}
where $\mathbf{w}_i$ is the learnable prototype of the $i$-th coarse class. We further use a softmax plus cross-entropy loss to supervise the similarity scores, which is formulated as:
\begin{equation}
    \mathcal{L}_\texttt{coarse}= - 
    log \frac{\mathbf{p}_y^\mathrm{T}\mathbf{w}_y}{\sum_{i=1}^M \exp\left(\mathbf{p}_{i}^\mathrm{T}\mathbf{w}_i\right)},
\end{equation}
where $y$ is the coarse label of the input image. We note there is a difference between the above coarse classification and the popular classification: the popular classification usually compares a single representation against a set of prototypes. In contrast, our coarse classification conducts a set-to-set comparison (\emph{i.e.}, $M$ tokens against $M$ prototypes).

\subsection{Overall Structure}
\textbf{Multiple transformer blocks for multi-level hierarchy.}
Some HIC datasets (\emph{e.g.}, ImageNet-1000) have a multi-level hierarchy. According to the coarse-to-fine multi-level hierarchy, TransHP may stack multiple prompting blocks. Each prompting block is responsible for a single level in the hierarchy, and the prompting block for the coarser level is placed closer to the bottom of the transformer. The detailed structure is shown in Appendix \ref{App: whole.}. Correspondingly, the overall loss function is formulated as: 
\begin{equation}
\mathcal{L} = \mathcal{L}_\texttt{fine} + \sum\nolimits_{l} \lambda_{l} \cdot \mathcal{L}^{l}_\texttt{coarse},
\end{equation}
where $\mathcal{L}_\texttt{fine}$ is the final classification loss, $\mathcal{L}^{l}_\texttt{coarse}$ is the coarse-level loss from the $l$-th prompting block, and $\lambda_{l}$ is the balance parameter.\par

Through the above training, the prompting layer explicitly learns the coarse-class prompts, as well as predicts the coarse class of the input image. 

\begin{table}[t]
\caption{The balance parameters used for $\mathcal{L}_{coarse}$ of different levels (The last $1$ is the balance parameter for the final classification.). ``-'' denotes that this transformer layer does not have prompt tokens.}
\vspace*{2mm}

\small
  \begin{tabularx}{\hsize}{l|Y|Y|Y|Y|Y|Y|Y|Y|Y|Y|Y|Y}
    \hline
    $\lambda$&$0$&$1$&$2$&$3$&$4$&$5$&$6$&$7$&$8$&$9$&$10$&$11$\\ \hline
ImageNet &$0.1$&$0.1$&$0.1$&$0.1$&$0.1$&$0.15$&$0.15$&$0.15$&$
0.15$&$1$&$1$&$1$\\
iNaturalist-2018 &$-$&$-$&$-$&$-$&$-$&$-$&$1$&$-$&$-$&$-$&$-$&$1$\\
iNaturalist-2019 &$-$&$-$&$-$&$-$&$-$&$-$&$1$&$-$&$-$&$-$&$-$&$1$\\
CIFAR-100&$-$&$-$&$-$&$-$&$-$&$-$&$-$&$-$&$1$&$-$&$-$&$1$\\
DeepFashion&$-$&$-$&$-$&$-$&$-$&$-$&$0.5$&$-$&$1$&$-$&$-$&$1$\\
 \hline

  \end{tabularx}
  \label{Table: lambda}
  \vspace*{-2mm}
  \\
\end{table}
\textbf{The position of the prompting block.} Currently, we do not have an exact position scheme for inserting the prompting block, given that the hierarchy of different datasets varies a lot. However, we recommend a qualitative principle to set the position for inserting the prompts: if the number of coarse classes is small (large), the position of the corresponding prompting blocks should be close to the bottom (top). Based on this principle, we can obtain the roughly-optimal position scheme through cross-verification. We empirically find that TransHP is robust to the position scheme to some extent (Fig.~\ref{Fig: level} in the Appendix). Table \ref{Table: lambda} summarizes the setting of the balance parameters and the position of prompting layers. 

\begin{figure*}[t] 
\centering 
\includegraphics[width=1\textwidth]{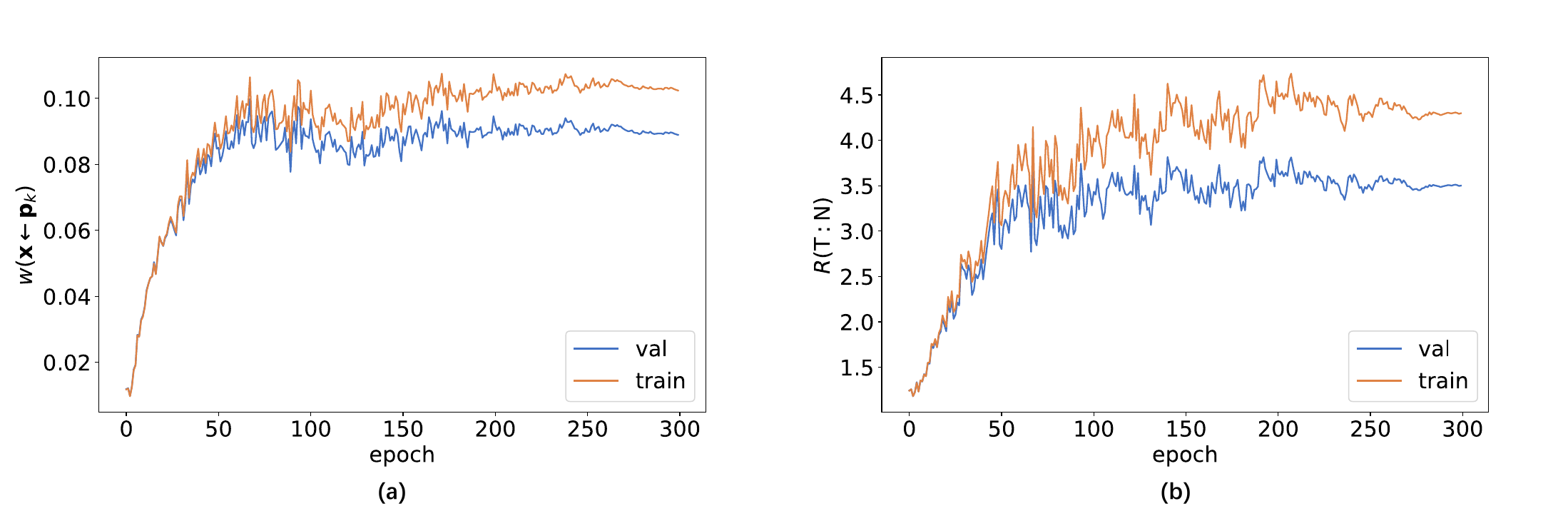} 
\caption{TransHP gradually focuses on the predicted coarse class when absorbing the prompts, yielding an autonomous selection. (a) The absorption weight of the target prompt. (b) The ratio of the target prompt weight against the largest non-target prompt weight. The dataset is CIFAR-100. We visualize these statistics on both the training and validation sets.}
\label{Fig: statistics} 
\end{figure*}
\subsection{TransHP Spontaneously Selects the Target Prompt} \label{Sec: selection}
We recall that we do not manually select the coarse-class prompt for TransHP. Instead, we concatenate the entire prompt set, \emph{i.e.}, $\mathbf{P}_M=\{\mathbf{p}_1, \mathbf{p}_2, \cdots, \mathbf{p}_M\}$, with the feature tokens. In this section, we will show that after TransHP is trained to convergence, the prompting block will spontaneously select the target prompt $\mathbf{p}_k$ ($k$ is the predicted coarse class) for absorption and largely neglect the non-target prompts $\mathbf{p}_{i \neq k}$.

Specifically, the self-attention in the transformer make each token absorb information from all the tokens (\emph{i.e.}, the feature tokens and the prompt tokens). 
In Eqn.~\ref{equal: TransHP}, given a feature token $\mathbf{x} \in [\mathbf{x}_{cls}, \mathbf{X}]$ (the superscript is omitted for simplicity), we derive its absorption weights on the $i$-th prompt token from the self-attention, which is formulated as:

\begin{equation}\label{equal: selection}
\small{w(\mathbf{x}\leftarrow \mathbf{p}_i) = \frac{\exp{(Q(\mathbf{x})^{\mathrm{T}} K(\mathbf{p}_i)/ \sqrt{d})}}{\sum \exp{(Q\Large(\mathbf{x})^{\top} K([\mathbf{x}_{cls}, \mathbf{X}, \mathbf{P}_M])\Large /\sqrt{d} ) }},}
\end{equation}
where $Q()$ and $K()$ project the input tokens into query and keys, respectively. $d$ is the scale factor. 

Based on the absorption weights, we consider two statistics:

$\bullet$ The absorption weight of the target prompt, \emph{i.e.}, $w(\mathbf{x}\leftarrow \mathbf{p}_k)$. It indicates the importance of the target prompt among all the tokens. 

$\bullet$ The absorption ratio between the target / largest non-target prompt, \emph{i.e.}, $R(\texttt{T:N}) = w(\mathbf{x}\leftarrow\mathbf{p}_k) / \max \{w(\mathbf{x}\leftarrow\mathbf{p}_{i \neq k})\}$. It measures the importance of the target prompt compared with the most prominent non-target prompt. 


Fig.~\ref{Fig: statistics} visualizes these statistics at each training epoch on CIFAR-100 \cite{krizhevsky2009learning}, from which we make two observations:

\textbf{Remark 1: The importance of the target prompt gradually increases to a high level.} From Fig.~\ref{Fig: statistics} (a), it is observed that the absorption weight on the prompt token undergoes a rapid increase and finally reaches about $0.09$.  We note that $0.09$ is significantly larger than the averaged weight 1/217 (1 class token + 196 patch tokens + 20 prompt tokens). 

\textbf{Remark 2: The target prompt gradually dominates among all the prompts.} 
From Fig.~\ref{Fig: statistics} (b), it is observed that the absorption weight on the target prompt 
gradually becomes much larger than the non-target prompt weight (about $4\times$). 

Combining the above two observations, we infer that during training, the prompting block of TransHP learns to focus on the target prompt $\mathbf{p}_k$ (within the entire prompt pool $\mathbf{P}_M$) for prompt absorption (Remark 2), yielding a soft-weighted selection on the target prompt. This dynamic absorption on the target prompt largely impacts the self-attention in the prompting layer (Remark 1) and conditions the subsequent feature extraction. 

\begin{figure*}[t] 
\centering 
\includegraphics[width=1\textwidth]{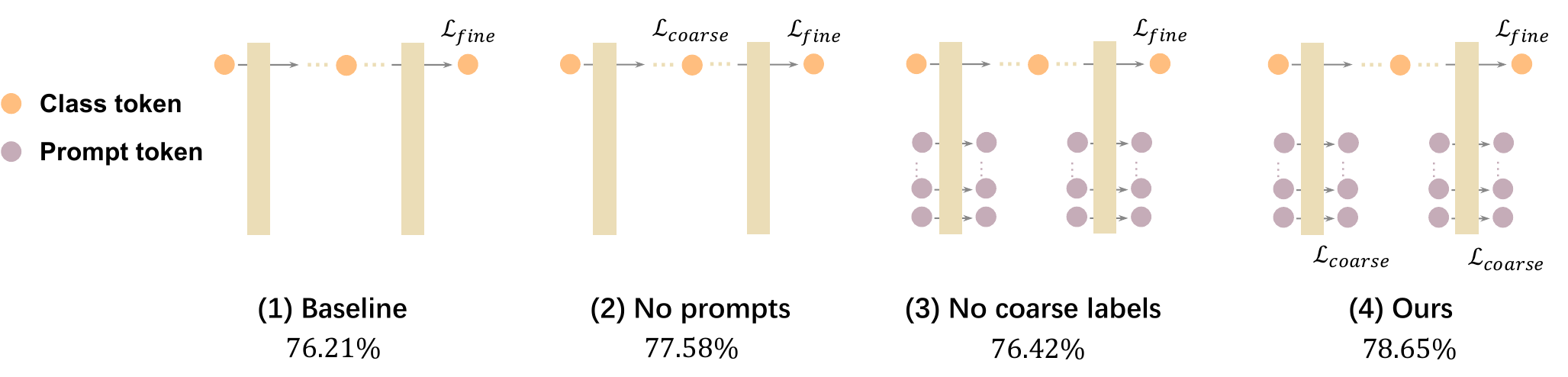} 
\caption{Comparison between TransHP and its variants on ImageNet. 1) A variant uses the coarse labels to supervise the class token in the intermediate layers (No prompts). 2) A variant injects additional tokens without supervision from the coarse-class labels (No coarse labels). 3) TransHP injects coarse-class information through prompt tokens and achieves the largest improvement (Ours). }
\vspace*{-2mm}
\label{Fig: cases} 
\end{figure*}

\begin{table*}[t]
\caption{The top-$1$ accuracy of TransHP on some other datasets (besides ImageNet). ``w Pre'' or ``w/o Pre'' denotes the models are trained from ImageNet pre-training or from scratch, respectively.}
\vspace*{2mm}
\small
  \begin{tabularx}{\hsize}{p{2.5cm}<{\centering}|Y|Y|Y|Y}
    \hline
Accuracy $\left( \% \right)  $ &iNaturalist-2018&iNaturalist-2019&CIFAR-100&DeepFashion
\\ \hline
Baseline (w/o Pre)&$51.07$& $57.33$&$61.77$&$83.42$\\
TransHP (w/o Pre)&$53.22$ &$59.24$ & $67.09$&$85.72$\\ \hline
Baseline (w Pre)& $63.01$ &$69.31$ &$84.98$ &$88.54$ \\
TransHP (w Pre)&$64.21$ &$71.62$ &$86.85$ & $89.93$\\
 \hline
  \end{tabularx}
  \vspace*{-4mm}
  \label{Table: other}
  \\
\end{table*}

Fig.\ref{Fig: main} (ii) further visualizes some instances from CIFAR-100 (20 coarse classes) for intuitively understanding the prompt absorption. 
We note that the coarse prediction may sometimes be incorrect. Therefore, we use the red (green) column to mark the prompts of the true (false) coarse class, respectively. In (a) and (b), TransHP correctly recognizes the coarse class of the input images and makes accurate prompt selection. The prompt of the true class  has the largest absorption weight and thus dominates the prompt absorption. In (c) and (d), TransHP encounters some confusion for distinguishing two similar coarse classes (due to their inherent similarity or image blur), and thus makes ambiguous selection.
In (e) and (f), TransHP makes incorrect coarse-class prediction and correspondingly selects the  prompt of a false class as the target prompt. 

\section{Experiments}
\begin{table*}[t]
\caption{TransHP brings consistent improvement on various backbones on ImageNet.}
\vspace*{2mm}
\small
  \begin{tabularx}{\hsize}{Y|Y|Y|Y|Y}
    \hline
Accuracy $\left( \% \right)  $&ViT-B/16&ViT-L/16&DeiT-S& DeiT-B\\ \hline
Baseline&$76.68^*$&$76.37^*$&$79.82$&$81.80$\\
TransHP&$79.51$&$78.80$&$80.55$&$82.35$\\

 \hline
\multicolumn{5}{l}{\small $*$ The performance of our reproduced ViT-B/16 and ViT-L/16 are slightly worse than 77.91 and 76.53 in its }\\
\multicolumn{5}{l}{original paper \cite{dosovitskiy2021an}, respectively.}\\

  \end{tabularx}
  \label{Table: backbones}
  \\
\vspace*{-2mm}
\end{table*}

\begin{table*}[t]
\caption{Comparison between TransHP and two most recent state-of-the-art methods. We replace their CNN backbones with the relatively strong transformer backbone for fair comparison.}
\vspace*{2mm}
\small
  \begin{tabularx}{\hsize}{Y|Y|Y|Y|Y|Y}
    \hline
Accuracy $\left( \% \right)  $ &ImageNet & iNat-2018&iNat-2019&CIFAR-100&DeepFashion
\\ \hline
Baseline&$76.21$&$63.01$ &$69.31$ &$84.98$ &$88.54$\\
Guided &$76.05$&$63.11$&$69.66$&$85.10$&$88.32$ \\
HiMulConE &$77.52$&$63.46$&$70.87$&$85.43$&$88.87$ \\ \hline
TransHP &$\textbf{78.65}$&$\textbf{64.21}$ &$\textbf{71.62}$ &$\textbf{86.85}$ & $\textbf{89.93}$\\
 \hline

  \end{tabularx}
  \label{Table: SOTA}
\end{table*}

\subsection{Implementation Details} 
\textbf{Datasets}. We evaluate the proposed TransHP on five  datasets with hierarchical labels, \emph{i.e.}, ImageNet \cite{deng2009imagenet}, iNaturalist-2018 \cite{van2018inaturalist}, iNaturalist-2019 \cite{van2018inaturalist}, CIFAR-100 \cite{krizhevsky2009learning}, and DeepFashion-inshop \cite{liu2016deepfashion}. The hierarchical labels of ImageNet are from WordNet \cite{miller1998wordnet}, with details illustrated on \href{https://observablehq.com/@mbostock/imagenet-hierarchy}{Mike's website}. Both the iNaturalist-2018/2019 have two-level hierarchical annotations: a super-category ($14$/$6$ classes) for the genus, and $8,142$/$1,010$ categories for the species. CIFAR-100 also has two-level hierarchical annotations: the coarse level has $20$ classes, and the fine level has $100$ classes. DeepFashion-inshop is a retrieval dataset with three-level hierarchy. To modify it for the  classification task, we random select $1/2$ images from each class for training, and the remaining $1/2$ images for validation. Both the training and validation set contain $2$ coarse classes, $17$ middle classes, and $7,982$ fine classes, respectively. 

\textbf{Training and inference details.} Our TransHP adopts an end-to-end training process. We use a lightweight transformer as our major baseline, which has $6$ heads (half of ViT-B) and $12$ blocks. The dimension of the embedding and the prompt token is $384$ (half of ViT-B). We train it for $300$ epochs on $8$ Nvidia A100 GPUs and PyTorch. The base learning rate is $0.001$ with cosine learning rate. We set the batch size, the weight decay and the number of warming up epochs as $1,024$, $0.05$ and $5$, respectively. Importantly, TransHP only adds small overhead to the baseline. Specifically, compared with the baseline ($22.05$ million parameters), our TransHP only adds $0.60$ million parameters (about $+2.7\%$) for ImageNet. When using ViT-B as the backbone, our TransHP only adds $+1.4\%$ parameters.
Due to the increase of parameters and the extra cost of the backward of several $L_{coarses}$, the training time increases by $15\%$ on our baseline and $12\%$ on ViT-B for ImageNet. For inference, the computation overhead is very light. The baseline and TransHP both use around $50$ seconds to finish the ImageNet validation with $8$ A100 GPUs.

\subsection{TransHP Improves the Accuracy}
\textbf{Improvement on ImageNet and the ablation study.}
We validate the effectiveness of TransHP on ImageNet and conduct the ablation study by comparing TransHP against two variants, as well as the baseline. As illustrated in Fig.~\ref{Fig: cases}, the two variants are: 1) we do not inject any prompts, but use the coarse labels to supervise the class token in the intermediate layers: similar with the final fine-level classification, the class token is also used for coarse-level classification. 2) we inject learnable tokens, but do not use the coarse labels as their supervision signal. Therefore, these tokens do not contain any coarse class information. From Fig.~\ref{Fig: cases}, we draw three observations as below: 
\textbf{1)} Comparing TransHP against the baseline, we observe a clear improvement of $+2.44\%$ top-1 accuracy, confirming the effectiveness of TransHP on ImageNet classification. \textbf{2)} Variant 1 (``No prompts'')  achieves some improvement ($+1.37\%$) over the baseline as well, but is still lower than TransHP by $-1.07\%$. It shows that using the hierarchical labels to supervise the intermediate state of the class token is also beneficial. However, since it does not absorb the prompting information, the improvement is relatively small. We thus infer that the hierarchical prompting is a superior approach for utilizing the hierarchical labels. \textbf{3)} Variant 2 (``No coarse labels'') barely achieves any improvement over the baseline, though it also increases the same amount of parameters as TransHP. It indicates that the benefit of TransHP is not due to the increase of some trainable tokens. Instead, the coarse class information injected through the prompt tokens matters.

\textbf{TransHP gains consistent improvements on more datasets.} Besides the most commonly used dataset ImageNet, we also conduct experiments on some other datasets, \emph{i.e.}, iNaturalist-2018, iNaturalist-2019, CIFAR-100 and DeepFashion. For these datasets, we use two settings, \emph{i.e.}, training from scratch (w/o Pre) and finetuning from the ImageNet-pretrained model (w Pre). The experimental results are shown in Table \ref{Table: other}, from which we draw two observations. First, under both settings, TransHP brings consistent improvement over the baselines. Second, when there is no pre-training, the improvement is even larger, especially on small datasets. For example, we note that on the smallest CIFAR-100, the improvement under ``w/o Pre'' and ``w Pre'' are $+5.32\%$ and $+1.87\%$, respectively. We infer it is because TransHP considerably alleviates the data-hungry problem of the transformer, which is further validated in Section \ref{sec: eff}.

\begin{table*}[t]
\caption{Comparison between TransHP and prior state-of-the-art hierarchical classification methods under the insufficient data scenario. ``$N\%$" means using $N\%$ ImageNet training data.}
\vspace*{2mm}
\small
  \begin{tabularx}{\hsize}{Y|Y|Y|Y|Y}
    \hline
Accuracy $\left( \% \right)  $&$100\%$&$50\%$&$20\%$&$10\%$\\ \hline
Baseline&$76.21$&$67.87$&$44.60$&$25.24$\\
Guided &$76.05$&$67.74$&$45.02$&$25.67$\\
HiMulConE &$77.52$&$69.23$&$48.50$&$30.76$ \\ \hline
TransHP&$\textbf{78.65}$&$\textbf{70.74}$&$\textbf{53.71}$&$\textbf{37.93}$\\

 \hline

  \end{tabularx}
  \label{Table: small}
  \\
  \vspace*{-4mm}
\end{table*}

\textbf{TransHP improves various backbones.} Besides the light transformer baseline used in all the other parts of this section, Table \ref{Table: backbones} evaluates the proposed TransHP on some more backbones, \emph{i.e.}, ViT-B/16 \cite{dosovitskiy2021an}, ViT-L/16 \cite{dosovitskiy2021an}, DeiT-S \cite{touvron2021training}, and DeiT-B \cite{touvron2021training}.
We observe that for the ImageNet classification, TransHP gains $2.83\%$, $2.43\%$, $0.73\%$, and $0.55\%$ improvement on these four backbones, respectively. \par 
\textbf{Comparison with state-of-the-art hierarchical classification methods.} We compare the proposed TransHP with two most recent hierarchy-based methods, \textit{i.e.} Guided \cite{landrieu2021leveraging}, HiMulConE \cite{zhang2022use}. 
We do not include more competing methods because most prior works are based on the convolutional backbones and are thus not directly comparable with ours. Since the experiments on large-scale datasets is very time-consuming, we only select the most recent state-of-the-art methods and re-implement them on the same transformer backbone (based on their released code). 
The experimental results are shown in Table \ref{Table: SOTA}.
It is clearly observed that the proposed TransHP achieves higher improvement and surpasses the two competing methods. 
For example, on the five datasets, TransHP surpasses the most recent state-of-the-art HiMulConE by $+1.13\%$ (ImageNet), $+0.75\%$ (iNat-2018), $+0.75\%$ (iNat-2019), $+1.42\%$ (CIFAR-100) and $1.06\%$ (DeepFashion), respectively. We also notice that while Guided achieves considerable improvement on the CNN backbones, its improvement over our transformer backbone is trivial. This is reasonable because improvement over higher baseline (\emph{i.e.}, the transformer backbone) is relatively difficult. This observation is consistent with \cite{zhang2022use}.

\subsection{TransHP Improves Data Efficiency}\label{sec: eff}
We investigate TransHP under the data-scarce scenario. To this end, we randomly select $1/10$, $1/5$, and $1/2$ training data from each class in ImageNet (while keeping the validation set untouched). The results are summarized in Table \ref{Table: small}, from which we draw three observations as below: 

\textbf{F}irst, as the training data decreases, all the methods undergo a significant accuracy drop. This is reasonable because the deep learning method in its nature is data-hungry, and arguably this data-hungry problem is further underlined in transformer \cite{dosovitskiy2021an}. \textbf{S}econd, compared with the baseline and two competing hierarchy-based methods, TransHP presents much higher resistance against the data decrease. For example, when the training data is reduced from 100\% $\rightarrow$ 10\%, the accuracy drop of the baseline and two competing methods are  $50.97\%$, $50.38\%$ and $46.76\%$, respectively. In contrast, the accuracy drop of the proposed TransHP ($40.72\%$) is significantly smaller. \textbf{T}hird, since TransHP undergoes relatively smaller accuracy decrease, its superiority under the low-data regime is even larger. For example, its surpasses the most competing HiMulConE by $1.13\%$, $1.51\%$, $5.21\%$ and $7.17\%$ under the 100\%, 50\%, 20\% and 10\% training data, respectively. Combining all these observations, we conclude that TransHP improves the data efficiency. The efficiency can be explained intuitively by drawing upon two perspectives, one philosophical and the other technical. \textbf{Philosophical Perspective:} Imagine knowledge as the essence of everything humans have summarized over time. When you possess knowledge, you have the distilled essence of myriad experiences and learnings. The proposed method leverages this accumulated knowledge. In scenarios where data is limited, the power of such distilled knowledge becomes even more pronounced. \textbf{Technical Perspective:} Now, think of data not just as isolated pieces of information but in categories. Even when the dataset might seem limited, there could still be ample samples within broader categories. This means that for these 'coarser' categories, accuracy can be achieved rapidly. Once the accuracy at this coarse level is established, the model can then use this foundation to prompt further. It's like planting a tree - you start with a strong base and then branch out.


\begin{figure*}[t] 
\centering 
\includegraphics[width=1\textwidth]{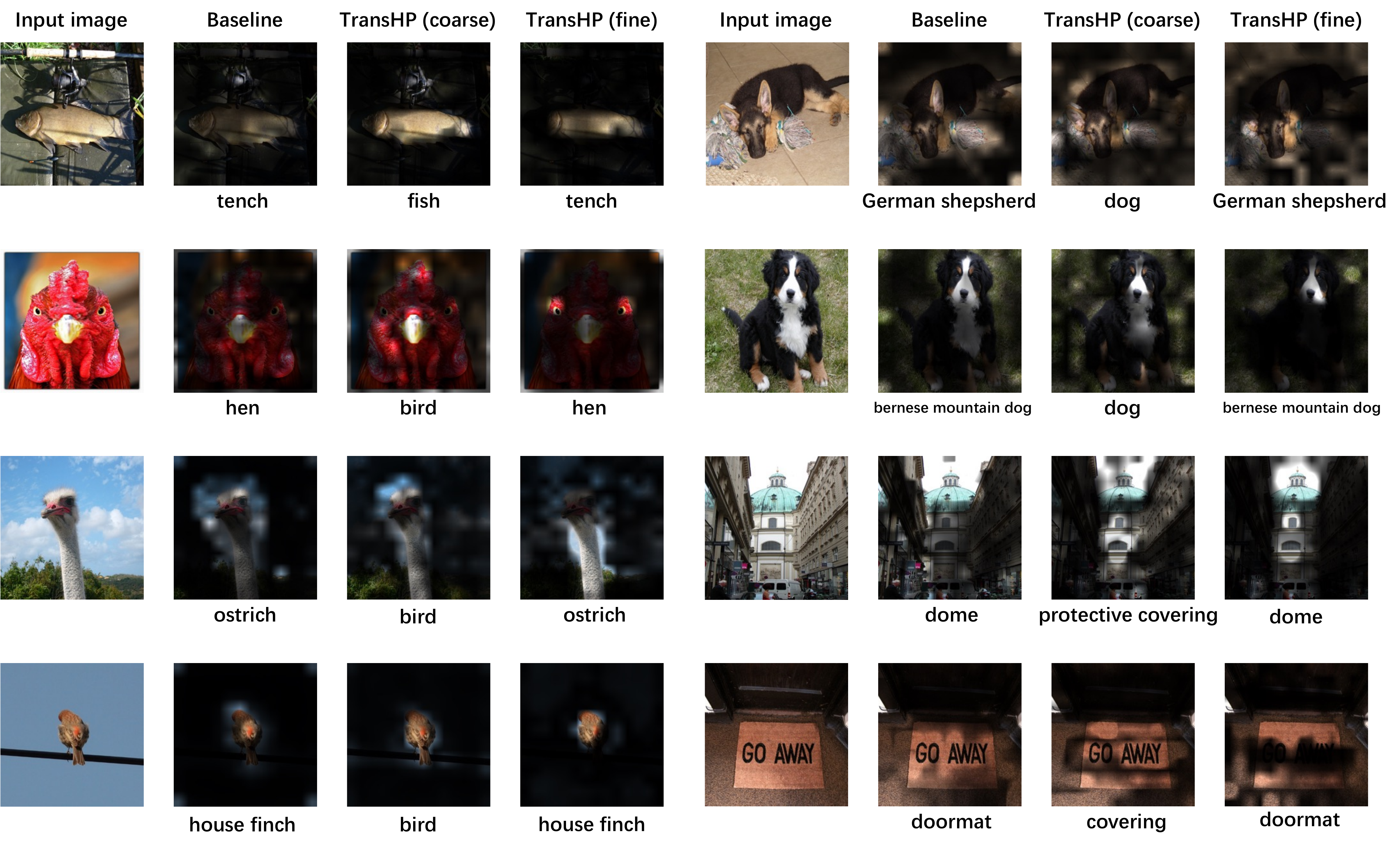} 
\caption{Visualization of the attention map for analyzing the receptive field. For TransHP, we visualize a block before and after receiving the prompt (\emph{i.e.}, coarse and fine), respectively. The ``coarse'' block favors an overview for coarse recognition, and the ``fine'' block further filters out the non-relevant regions after receiving the prompt.} 
\vspace{-1mm}
\label{Fig: subtle} 
\end{figure*}
\subsection{TransHP Improves Model Explainability}\label{sec: visualization}
We analyze the receptive field of the class token to understand how TransHP reaches its prediction. 
Basically, the transformer integrates information across the entire image according to the attention map, yielding its receptive field. Therefore, we visualize the attention map of the class token in Fig.~\ref{Fig: subtle}. For the proposed TransHP, we visualize the attention map at the prompting block (which handles the coarse-class information) and the last block (which handles the fine-class information). For the ViT baseline, we only visualize the attention score map of the the last block. We draw two observations from Fig.~\ref{Fig: subtle}:

First, TransHP has a different attention pattern compared with the baseline. The baseline attention generally covers the entire foreground, which is consistent with the observation in \cite{dosovitskiy2021an}. In contrast, in TransHP, although the coarse block attends to the overall foreground as well, the fine block concentrates its attention on relatively small and critical regions, in pace with the ``prompting $\rightarrow$ predicting'' procedure. For example, given the ``hen'' image on the second row (left), TransHP attends to the overall foreground before receiving the coarse-class prompt (\emph{i.e.}, the bird) and focuses to the eyes and bill for recognizing the ``hen'' out from the ``bird''.
Second, TransHP shows better capacity for ignoring the redundant and non-relevant regions. For example, given the ``doormat'' image on the fourth row (right), TransHP ignores the decoration of ``GO AWAY'' after receiving the coarse-class prompt of ``covering''. Similar observation is with the third row (right), where TransHP ignores the walls when recognizing the ``dome'' out from ``protective covering''.
\vspace{-1mm}
\section{Conclusion}
This paper proposes a novel Transformer with Hierarchical Prompting (TransHP) for image classification. 
Before giving its final prediction, TransHP predicts the coarse class with an intermediate layer and correspondingly injects the coarse-class prompt to condition the subsequent inference. An intuitive effect of our hierarchical prompting is: TransHP favors an overview of the object for coarse prediction and then concentrates its attention to some critical local regions after receiving the prompt, which is similar to the human visual recognition.
We validate the effectiveness of TransHP through extensive experiments and hope the hierarchical prompting reveals a new 
insight for understanding the transformers.

\textbf{Limitation.} We presently focus on the image classification task in this paper, while there are some other tasks that are the potential to benefit from hierarchical annotations, e.g., semantic segmentation. Therefore, we would like to extend TransHP for more visual recognition tasks in the future. 

\bibliography{nips2023_conference}

\begin{thebibliography}{10}

\bibitem{miller1998wordnet}
George~A Miller.
\newblock {\em WordNet: An electronic lexical database}.
\newblock MIT press, 1998.

\bibitem{mikolov2013efficient}
Tomas Mikolov, Kai Chen, Greg Corrado, and Jeffrey Dean.
\newblock Efficient estimation of word representations in vector space.
\newblock In {\em International Conference on Learning Representations
  Workshop}, 2013.

\bibitem{landrieu2021leveraging}
Loic Landrieu and Vivien Sainte~Fare Garnot.
\newblock Leveraging class hierarchies with metric-guided prototype learning.
\newblock In {\em British Machine Vision Conference (BMVC)}, 2021.

\bibitem{zhang2022use}
Shu Zhang, Ran Xu, Caiming Xiong, and Chetan Ramaiah.
\newblock Use all the labels: A hierarchical multi-label contrastive learning
  framework.
\newblock In {\em Proceedings of the IEEE/CVF Conference on Computer Vision and
  Pattern Recognition}, pages 16660--16669, 2022.

\bibitem{li2021prefix}
Xiang~Lisa Li and Percy Liang.
\newblock Prefix-tuning: Optimizing continuous prompts for generation.
\newblock {\em arXiv preprint arXiv:2101.00190}, 2021.

\bibitem{gu2021ppt}
Yuxian Gu, Xu~Han, Zhiyuan Liu, and Minlie Huang.
\newblock Ppt: Pre-trained prompt tuning for few-shot learning.
\newblock {\em arXiv preprint arXiv:2109.04332}, 2021.

\bibitem{he2021towards}
Junxian He, Chunting Zhou, Xuezhe Ma, Taylor Berg-Kirkpatrick, and Graham
  Neubig.
\newblock Towards a unified view of parameter-efficient transfer learning.
\newblock {\em arXiv preprint arXiv:2110.04366}, 2021.

\bibitem{ge2022domain}
Chunjiang Ge, Rui Huang, Mixue Xie, Zihang Lai, Shiji Song, Shuang Li, and Gao
  Huang.
\newblock Domain adaptation via prompt learning.
\newblock {\em arXiv preprint arXiv:2202.06687}, 2022.

\bibitem{dosovitskiy2021an}
Alexey Dosovitskiy, Lucas Beyer, Alexander Kolesnikov, Dirk Weissenborn,
  Xiaohua Zhai, Thomas Unterthiner, Mostafa Dehghani, Matthias Minderer, Georg
  Heigold, Sylvain Gelly, Jakob Uszkoreit, and Neil Houlsby.
\newblock An image is worth 16x16 words: Transformers for image recognition at
  scale.
\newblock In {\em International Conference on Learning Representations}, 2021.

\bibitem{deng2009imagenet}
Jia Deng, Wei Dong, Richard Socher, Li-Jia Li, Kai Li, and Li~Fei-Fei.
\newblock Imagenet: A large-scale hierarchical image database.
\newblock In {\em 2009 IEEE conference on computer vision and pattern
  recognition}, pages 248--255. Ieee, 2009.

\bibitem{van2018inaturalist}
Grant Van~Horn, Oisin Mac~Aodha, Yang Song, Yin Cui, Chen Sun, Alex Shepard,
  Hartwig Adam, Pietro Perona, and Serge Belongie.
\newblock The inaturalist species classification and detection dataset.
\newblock In {\em Proceedings of the IEEE conference on computer vision and
  pattern recognition}, pages 8769--8778, 2018.

\bibitem{johannesson2016visual}
Omar~I Johannesson, Ian~M Thornton, Irene~J Smith, Andrey Chetverikov, and Arni
  Kristj{\'a}nsson.
\newblock Visual foraging with fingers and eye gaze.
\newblock {\em i-Perception}, 7(2):2041669516637279, 2016.

\bibitem{mirza2018human}
M~Berk Mirza, Rick~A Adams, Christoph Mathys, and Karl~J Friston.
\newblock Human visual exploration reduces uncertainty about the sensed world.
\newblock {\em PloS one}, 13(1):e0190429, 2018.

\bibitem{wang2022learning}
Zifeng Wang, Zizhao Zhang, Chen-Yu Lee, Han Zhang, Ruoxi Sun, Xiaoqi Ren,
  Guolong Su, Vincent Perot, Jennifer Dy, and Tomas Pfister.
\newblock Learning to prompt for continual learning.
\newblock In {\em Proceedings of the IEEE/CVF Conference on Computer Vision and
  Pattern Recognition}, pages 139--149, 2022.

\bibitem{wang2022dualprompt}
Zifeng Wang, Zizhao Zhang, Sayna Ebrahimi, Ruoxi Sun, Han Zhang, Chen-Yu Lee,
  Xiaoqi Ren, Guolong Su, Vincent Perot, Jennifer Dy, et~al.
\newblock Dualprompt: Complementary prompting for rehearsal-free continual
  learning.
\newblock {\em European Conference on Computer Vision}, 2022.

\bibitem{luddecke2022image}
Timo L{\"u}ddecke and Alexander Ecker.
\newblock Image segmentation using text and image prompts.
\newblock In {\em Proceedings of the IEEE/CVF Conference on Computer Vision and
  Pattern Recognition}, pages 7086--7096, 2022.

\bibitem{zhang2022NOAH}
Yuanhan Zhang, Kaiyang Zhou, and Ziwei Liu.
\newblock Neural prompt search.
\newblock 2022.

\bibitem{jia2022vpt}
Menglin Jia, Luming Tang, Bor-Chun Chen, Claire Cardie, Serge Belongie, Bharath
  Hariharan, and Ser-Nam Lim.
\newblock Visual prompt tuning.
\newblock In {\em European Conference on Computer Vision (ECCV)}, 2022.

\bibitem{su2021semi}
Jong-Chyi Su and Subhransu Maji.
\newblock Semi-supervised learning with taxonomic labels.
\newblock {\em arXiv preprint arXiv:2111.11595}, 2021.

\bibitem{jain2023test}
Kanishk Jain, Shyamgopal Karthik, and Vineet Gandhi.
\newblock Test-time amendment with a coarse classifier for fine-grained
  classification.
\newblock {\em arXiv preprint arXiv:2302.00368}, 2023.

\bibitem{yan2015hd}
Zhicheng Yan, Hao Zhang, Robinson Piramuthu, Vignesh Jagadeesh, Dennis DeCoste,
  Wei Di, and Yizhou Yu.
\newblock Hd-cnn: hierarchical deep convolutional neural networks for large
  scale visual recognition.
\newblock In {\em Proceedings of the IEEE international conference on computer
  vision}, pages 2740--2748, 2015.

\bibitem{brown2020language}
Tom Brown, Benjamin Mann, Nick Ryder, Melanie Subbiah, Jared~D Kaplan, Prafulla
  Dhariwal, Arvind Neelakantan, Pranav Shyam, Girish Sastry, Amanda Askell,
  et~al.
\newblock Language models are few-shot learners.
\newblock {\em Advances in neural information processing systems},
  33:1877--1901, 2020.

\bibitem{gao2021making}
Tianyu Gao, Adam Fisch, and Danqi Chen.
\newblock Making pre-trained language models better few-shot learners.
\newblock In {\em ACL/IJCNLP (1)}, 2021.

\bibitem{jiang2020can}
Zhengbao Jiang, Frank~F Xu, Jun Araki, and Graham Neubig.
\newblock How can we know what language models know?
\newblock {\em Transactions of the Association for Computational Linguistics},
  8:423--438, 2020.

\bibitem{krizhevsky2009learning}
Alex Krizhevsky, Geoffrey Hinton, et~al.
\newblock Learning multiple layers of features from tiny images.
\newblock 2009.

\bibitem{liu2016deepfashion}
Ziwei Liu, Ping Luo, Shi Qiu, Xiaogang Wang, and Xiaoou Tang.
\newblock Deepfashion: Powering robust clothes recognition and retrieval with
  rich annotations.
\newblock In {\em Proceedings of IEEE Conference on Computer Vision and Pattern
  Recognition (CVPR)}, 2016.

\bibitem{touvron2021training}
Hugo Touvron, Matthieu Cord, Matthijs Douze, Francisco Massa, Alexandre
  Sablayrolles, and Herv{\'e} J{\'e}gou.
\newblock Training data-efficient image transformers \& distillation through
  attention.
\newblock In {\em International Conference on Machine Learning}, pages
  10347--10357. PMLR, 2021.

\end{thebibliography}
\bibliographystyle{unsrt}

\newpage
\appendix
\onecolumn
\section*{Appendix}

\section{Multiple layers of hierarchy}\label{App: whole.}
\begin{figure}[t] 
\centering 
\includegraphics[width=0.6\textwidth]{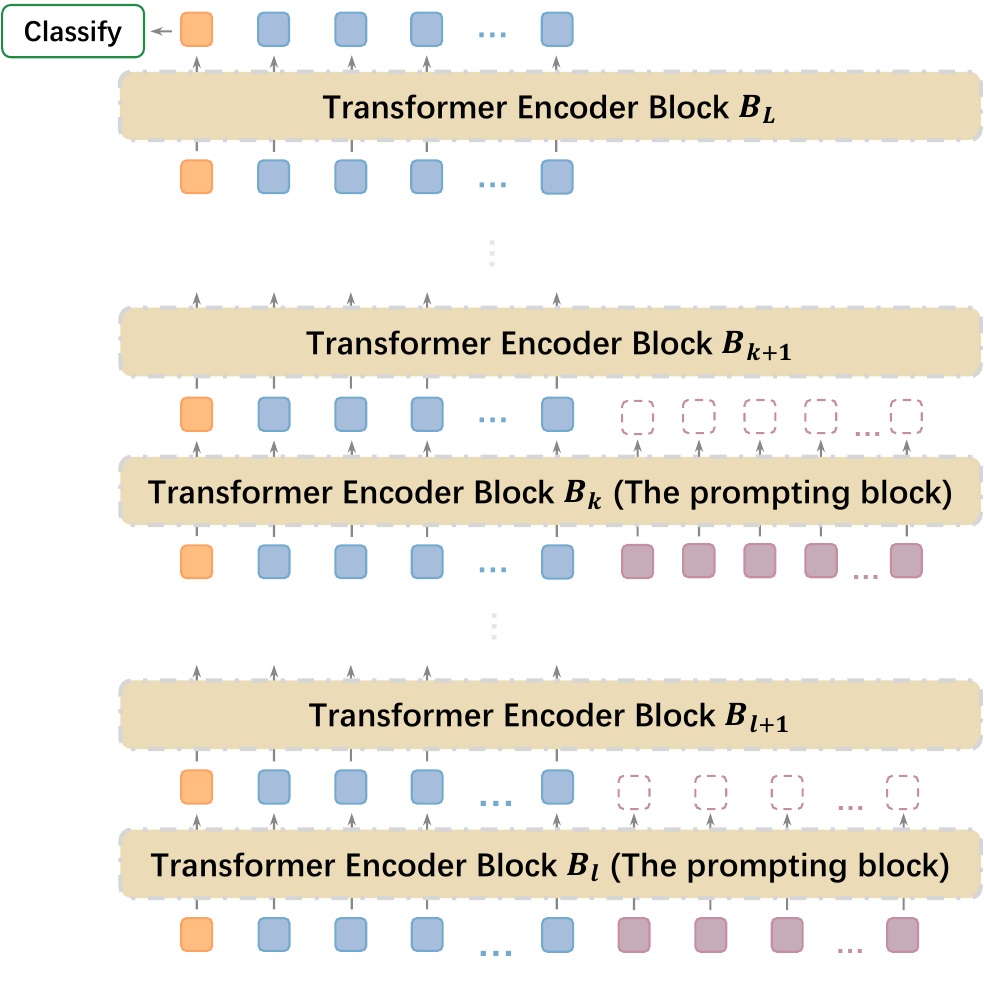} 
\caption{The illustration of TransHP with multiple layers of hierarchy. $k$ and $l$ are two insider layers, and $L$ is the final layer.} 
\label{Fig: multi} 
\end{figure}

We illustrate the TransHP in Fig. \ref{Fig: multi} when a dataset has multiple layers of hierarchy. 

\section{Coarse-level classes of CIFAR-100}\label{App: Coarse-level classes of CIFAR-100}
$[0]$: aquatic mammals, $[1]$: fish, $[2]$: flowers, $[3]$: food containers, $[4]$: fruit and vegetables, $[5]$: household electrical devices, $[6]$: household furniture, $[7]$: insects, $[8]$: large carnivores, $[9]$: large man-made outdoor things, $[10]$: large natural outdoor scenes, $[11]$: large omnivores and herbivores, $[12]$: medium mammals, $[13]$: non-insect invertebrates, $[14]$: people, $[15]$: reptiles, $[16]$: small mammals, $[17]$: trees, $[18]$: vehicles-1, and $[19]$: vehicles-2.



\begin{figure*}[t] 
\centering 
\includegraphics[width=0.6\textwidth]{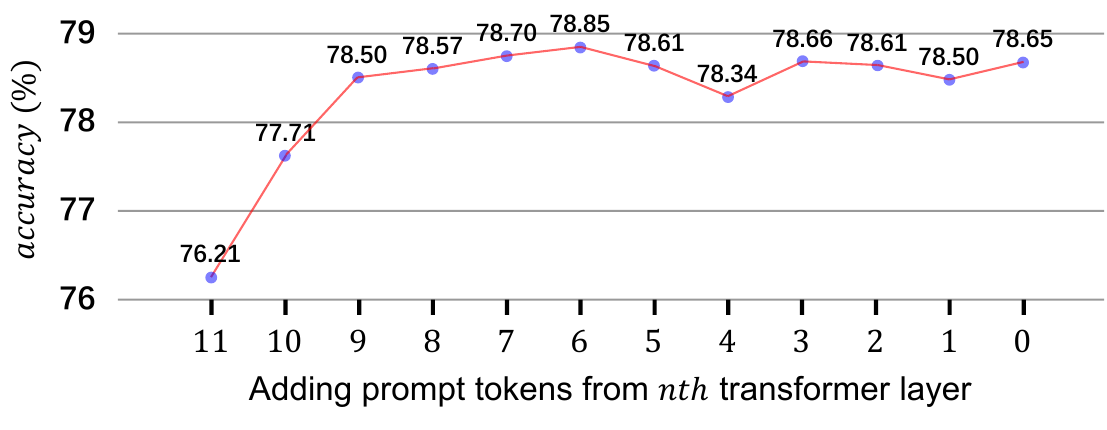} 
\caption{The top-$1$ accuracy on ImageNet \textit{w.r.t} the transformer layer from which to add prompt tokens. The highest two transformer layers (which do not1 have too coarse-level labels) play an important role.} 
\label{Fig: level} 
\end{figure*}
\begin{table*}[t]
\caption{The analysis of the number of coarse-level classes on the CIFAR-100 dataset. ``$N$-class" denotes that there are $N$ classes for the coarse-level classification.}
\vspace*{2mm}
\small
  \begin{tabularx}{\hsize}{l|Y|Y|Y|Y|Y}
    \hline
    Accuracy ($\%$)&baseline&2-class&5-class&10-class&20-class\\ \hline
w/o Pre &$61.77$&$63.34$&$63.12$&$64.47$&$67.09$\\
w Pre &$84.98$&$86.40$&$86.35$&$86.50$&$86.85$\\
 \hline

  \end{tabularx}
  \label{Table: label-abla}
  \\
\end{table*}

\section{Importance analysis of classification at different hierarchical levels}\label{App: Importance analysis of classification at different hierarchical levels.}

From Table \ref{Table: lambda} (Line $1$), each transformer layer is responsible for one level classification. We remove the prompt tokens from the coarsest level to the finest level. 
In Fig. \ref{Fig: level}, $n$ denotes that the prompt tokens are added from the $n$th transformer layer. We conclude that only the last two coarse level classifications  (arranged at the $9$th and $10$th transformer layer)  contribute most to the final classification accuracy. That means: (1) it is not necessary that the number of hierarchy and transformer layers are equal. (2) it is no need to adjust any parameters from too coarse level hierarchy. (Note that: though the current balance parameter for the $8$th transformer layer is $0.15$, when it is enlarged to $1$, no further improvement is achieved.)

\section{Analysis of the number of coarse-level classes}
As shown in Appendix  \ref{App: Coarse-level classes of CIFAR-100}, the CIFAR-100 dataset has 20 coarse-level classes.  When we combine them into 10 coarse-level classes, we have ([0-1]), ([2-17]), ([3-4]), ([5-6]), ([12-16]), ([8-11]), ([14-15]), ([9-10]), ([7-13]), and ([18-19]). When we combine them into 5 coarse-level classes, we have ([0-1-12-16]), ([2-17-3-4]), ([5-6-9-10]), ([8-11-18-19]), and ([7-13-14-15]). When we combine them into 2 coarse-level classes, we have ([0-1-7-8-11-12-13-14-15-16]) and ([2-3-4-5-6-9-10-17-18-19]). The experimental results are listed in Table \ref{Table: label-abla}.

\begin{table}[t]
\caption{Comparison between TransHP with the original baseline and the ``No prompts" baseline.}
\vspace*{2mm}
\small
  \begin{tabularx}{\hsize}{l|Y|Y|Y|Y}
    \hline
Accuracy $\left( \% \right)  $  & iNat-2018&iNat-2019&CIFAR-100&DeepFashion
\\ \hline
Baseline (w/o Pre)&$51.07$ &$57.33$ &$61.77$ &$83.42$\\
No prompts (w/o Pre)&$51.88$&$58.45$&$63.78$&$84.23$ \\
TransHP (w/o Pre) &$\textbf{53.22}$ &$\textbf{59.24}$ &$\textbf{67.09}$ & $\textbf{85.72}$\\\hline
Baseline (w Pre)&$63.01$ &$69.31$ &$84.98$ &$88.54$\\
No prompts (w Pre)&$63.41$&$70.73$&$85.50$&$89.59$ \\
TransHP (w Pre) &$\textbf{64.21}$ &$\textbf{71.62}$ &$\textbf{86.85}$ & $\textbf{89.93}$\\
 \hline

  \end{tabularx}
  \label{Table: no prompt}
    \vspace*{-2mm}
\end{table}

We observe that: 1) Generally, using more coarse-level classes is better. 2) Using only 2 coarse-level classes still brings over $1\%$ accuracy improvement.

\section{The comparison with the ``No prompts" baseline}\label{App: No prompt.}

In this section, we provide more experiments with the ``No prompts" baseline. The detail of the ``No prompts" baseline is shown in Fig. \ref{Fig: cases} (2). The experimental results are shown in Table \ref{Table: no prompt}. We find that though ``No prompts" baseline surpasses the original baseline, our TransHP still shows significant superiority over this baseline.

\section{Additional $L_{coarse}$ with DeiT.}\label{App: additional}
We introduce the experimental results by only adopting $L_{coarse}$ in DeiT. Note that the $L_{coarse}$ is imposed on the class token as shown in Fig. \ref{Fig: cases} (2). We find that the TransHP still shows performance improvement compared with only using $L_{coarse}$ on DeiT-S and DeiT-B: compared with DeiT-S ($79.82\%$) and DeiT-B ($81.80\%$), ``only with $L_{coarse}$" achieves $79.98\%$ and $81.76\%$ while the TransHP achieves $80.55\%$ and $82.35\%$, respectively.



\end{document}